\newif\ifBLIND
\newif\ifNOTBLIND
\BLINDfalse
\NOTBLINDtrue

\newif\ifUSENIPS
\newif\ifUSEICML
\newif\ifNOSTYLE
\USENIPStrue
\USEICMLfalse
\NOSTYLEfalse

\documentclass{article}

\ifNOSTYLE
    \usepackage[margin=1.4in]{geometry}
\fi
\usepackage[utf8]{inputenc}
\usepackage{amsmath}
\usepackage{amsthm}
\usepackage[hyperfootnotes=false]{hyperref}
\usepackage{amssymb}
\allowdisplaybreaks
\usepackage{graphicx}
\usepackage{float}
\usepackage{microtype}
\usepackage{footnote}
\usepackage{multirow}
\usepackage{physics}
\usepackage{footmisc}
\usepackage{array}
\usepackage{makecell}
\usepackage{booktabs}
\usepackage[english]{babel}
\usepackage[autostyle, english = american]{csquotes}
\MakeOuterQuote{"}
\usepackage{wrapfig}
\usepackage{subfig}

\ifUSENIPS
    \ifBLIND
        \usepackage{NIPS/nips_2018}
    \fi
    \ifNOTBLIND
        \usepackage[preprint]{NIPS/nips_2018}
    \fi
\fi

\ifUSEICML
    \usepackage{ICML/icml2018}
\else
    \usepackage{algorithm}
    \usepackage{algorithmic}
\fi

\usepackage{natbib}

\begin{document}

\ifUSEICML
\else
    \title{Natural Gradient Deep Q-learning}
    \author{Ethan Knight
    \\ \texttt{\href{mailto:ethan.h.knight@gmail.com}{ethan.h.knight@gmail.com}} \\Stanford Cognitive \& Systems Neuroscience Lab \\ The Nueva School
    \And
    Osher Lerner \\\ \texttt{\href{mailto:osherler@gmail.com}{osherler@gmail.com}} \\ The Nueva School
    }
    \maketitle
\fi

\begin{abstract}
    We present a novel algorithm to train a deep Q-learning agent using natural-gradient techniques. We compare the original deep Q-network (DQN) algorithm to its natural-gradient counterpart, which we refer to as NGDQN, on a collection of classic control domains. Without employing target networks, NGDQN significantly outperforms DQN without target networks, and performs no worse than DQN with target networks, suggesting that NGDQN stabilizes training and can help reduce the need for additional hyperparameter tuning. We also find that NGDQN is less sensitive to hyperparameter optimization relative to DQN. Together these results suggest that natural-gradient techniques can improve value-function optimization in deep reinforcement learning.
    
\end{abstract}

\section{Introduction}
A core piece of various reinforcement-learning algorithms is the estimation of a value function-- the expected sum of future discounted rewards under a desired policy \citep{sutton}. Q-learning is a model-free reinforcement learning algorithm that estimates the value function under the optimal policy by minimizing the temporal-difference error between the agent's value function estimates \citep{OG-qlearning}. This basic algorithm, when combined with deep neural networks \citep{lecun2015deep}, has proven to be a major success in AI, most notably by exhibiting human-level performance in a suite of challenging Atari games \citep{Mnih}.

Because Q-learning, as the control extension of TD algorithms \citep{sutton1988learning}, is not truly a stochastic gradient descent algorithm \citep{maei2011gradient}, convergence of the algorithm with non-linear function approximators is poorly understood. In fact, it has been shown that TD algorithms can sometimes be divergent \citep{tsitsiklis1997analysis}. Moreover, in practice it is sometimes difficult to train a deep neural network with Q-learning. In the original DQN work, for example, the authors proposed three key additions to stabilize training, namely experience replay \citep{lin1992self}, reward clipping, and the use of target networks. \citep{Mnih}

In this work, we aim to address some of the practical issues pertaining to DQN training as well as improve upon it by using natural-gradient techniques. Natural gradient was originally proposed by \citeauthor{Amari} as a method to accelerate gradient descent \citeyearpar{Amari}. Rather than exclusively using the loss gradient or using local curvature from the Hessian matrix, natural gradient uses "information" found in the parameter space of the model to train efficiently.

Natural gradient has been successfully applied to several deep learning domains \citep{Desjardins,trpo,acktr} and has been used to accelerate the training of reinforcement learning systems \citep{Kakade,peters2008natural,dabney2014natural}. To motivate our approach, we hoped that using natural gradient would accelerate the training of DQN, making our system more sample-efficient, thereby addressing one of the major problems in reinforcement learning. We also hoped that since natural gradient stabilizes training (e.g. natural gradient is relatively unchanged when changing the order of training inputs \citep{Pascanu}), NGDQN could be able to achieve good results without a target network, and converge to good solutions with more stability.

In our experiments, we observed both effects. When training without a target network, NGDQN converged much faster and more frequently than our DQN baseline, and its training appeared much more stable. Further, NGDQN performed no worse than DQN with target networks trained over the same number of episodes.

This paper was inspired by the Requests for Research list published by \citeauthor{r4r}, which has listed the application of natural-gradient techniques to Q-learning since June 2016 \citeyearpar{r4rgithub, r4r}. This paper presents the first successful attempt to our knowledge: our method to accelerate the training of Q-networks using natural gradient.

\section{Background}
\subsection{Reinforcement learning problem}
In the reinforcement learning problem, typically modeled as an MDP \citep{puterman2014markov}, an agent interacts with an environment to maximize cumulative reward. The agent observes a state $s$, performs an action $a$, and receives a new state $s'$ and reward $r$. Usually a discount factor $\gamma$ is also defined, which specifies the relative importance of immediate reward as opposed to those recieved in the future. More specifically the objective is to maximize:
\begin{equation}
E_{\pi}[R_t] = E\big[\sum_{t'=t}^T \gamma^{t'-t}r_{t'}|\pi\big]\ ,
\end{equation}
by attempting to learn a good policy $\pi$.

\subsection{Q-learning}
Q-learning \citep{OG-qlearning,OG-deepqlearning} is a model-free reinforcement learning algorithm which works by gradually learning $Q(s,a)$, the expectation of the cumulative reward. The Bellman equation defines the optimal Q-value $Q^*$ \citep{sutton,dqndemo}:
\begin{equation}
    Q^*(s, a) = \mathbb{E}\left[R(s, a) + \gamma \sum_{s'} P(s'|s, a) \max_{a'} Q^*(s', a') \right]
\end{equation}

This function $Q$ can be then optimized through value iteration, which defines the update rule $Q(s, a) \leftarrow \mathbb{E}\left[r+\gamma\max_{a'}Q(s',a')|s, a\right]$ \citep{sutton,Mnih}. Additionally, the optimal policy $\pi$ is defined as $\pi(s)=\text{argmax}_{a}Q^*(s,a)$ \citep{sutton,dqndemo}.
A neural network can be described as a parametric function approximator that uses "layers" of units, each containing weights, biases and activation functions, called "neurons". Each layer's output is fed into the next layer, and the loss is backpropagated to each layer's weights in order to adjust the parameters according to their effect on the loss.

For deep Q-learning, the neural network, parameterized by $\theta$, takes in a state $s$ and outputs a predicted future reward for each possible action $a$ with a linear activation on the final layer. The loss of this network is defined as follows, given the environment $\varepsilon$:

\begin{equation}
\label{qnetworkloss}
\mathcal{L} = \mathbb{E}\left[(y-Q(s, a_i; \theta))^2\right]
\end{equation}

where $Q(s, a_i; \theta)$ is the output of the network corresponding to action taken $a_i$, and
\begin{equation}
y = \mathbb{E}_{s'\sim\varepsilon}\left[r+\gamma\max_{a'}Q(s', a'; \theta) \big|s, a_i\right]
\end{equation}

Notice that we take the mean-squared-error between the expected Q-value and actual Q-value. The neural network is optimized over the course of numerous iterations through some form of gradient descent. In the original DQN (deep Q-network) paper in which an agent successfully played Atari games from pixels, an adaptive gradient method was used to train this network \citep{Mnih}. 

Deep Q-networks use experience replay to train the Q-value estimator on a randomly sampled batch of previous experiences (essentially replaying past remembered events back into the neural network) \citep{lin1992self}. Experience replay makes the training samples independent and identically distributed (i.i.d.), unlike the highly correlated consecutive samples which are encountered during interaction with the environment \citep{prioritized}. This is a prerequisite for many SGD convergence theorems. Additionally, DQN uses an $\epsilon$-greedy policy: the agent acts nearly randomly in order to explore potentially successful strategies, and as the agent learns, it acts randomly less often (this is sometimes called the "exploit" stage, as opposed to the prior "explore" stage). Mathematically, the probability of choosing a random action $\epsilon$ is gradually annealed over the course of training.

We combine these two approaches, using natural gradient to optimize the neural network in Q-learning architectures.

\section{Natural gradient for Q-learning}
Gradient descent optimizes parameters of a model with respect to a loss function by "descending" down the loss manifold. To do this, we take the gradient of the loss with respect to the parameters, then move in the opposite direction of that gradient \citep{Goodfellow}. Mathematically, gradient descent updates parameters $\theta$ of a model mapping from $x$ to $y$ as $\theta \gets \theta-\alpha \nabla_\theta \mathcal{L}(x, y; \theta)$  given a learning rate of $\alpha$.

A commonly used variant of gradient descent is stochastic gradient descent (SGD). Instead of calculating the entire gradient at a time, SGD uses a mini-batch of training samples: $\theta-\alpha \nabla_\theta \mathcal{L}(x_i, y_i; \theta)$. Our baselines use Adam, an adaptive gradient optimizer, which is a modification of SGD \citep{adam}.

However, this approach of gradient descent has a number of issues. For one, gradient descent will often become very slow in plateaus where the magnitude of the gradient is close to zero. Also, while gradient descent takes uniform steps in the parameter space, this does not necessarily correspond to uniform steps in the output distribution. Natural gradient attempts to fix these issues by incorporating the inverse Fisher information matrix, a concept from statistical learning theory \citep{Amari}.

Essentially, the core problem is that Euclidean distances in the parameter space do not give enough information about the distances between the corresponding outputs, as there is not a strong enough relationship between the two \citep{Foti}. \citeauthor{Kullback} define a more expressive distribution-wise measure, as follows \citeyearpar{Kullback}:
\begin{equation}
KL(\mu_1|\mu_2)=\int _{-\infty }^{\infty }\mu_1(s)\,\log {\frac {\mu_1(s)}{\mu_2(s)}}\,{\rm {d}}s\!
\end{equation}

However, since $KL(\mu_1|\mu_2)\neq KL(\mu_2|\mu_1)$, symmetric KL divergence, also known as Jensen-Shannon (JS) divergence, is defined as follows \citep{Foti}: 
\begin{equation}
KL_{\text{sym}}(\mu_1|\mu_2) := \frac{1}{2}\left(KL(\mu_1|\mu_2) +KL(\mu_2|\mu_1)\right)
\end{equation}

To perform gradient descent on the manifold of functions given by our model, we use the Fisher information metric on a Riemannian manifold. Since symmetric KL divergence behaves like a distance measure in infinitesimal form, a Riemannian metric is derived as the Hessian of the divergence of symmetric KL divergence \citep{Pascanu}. We give \citeauthor{Pascanu}'s definition, which assume that the probability of a point sampled from the network is a gaussian with the network's output as the mean and with a fixed variance. Given some probability density function $p$, input vector $s$, and parameters $\theta$ \citep{Pascanu}:
\begin{equation}
\mathbf{F}_\theta = \mathbb{E}_{s,q}[(\nabla\log p_\theta(q|s))^T(\nabla\log p_\theta(q|s))]
\end{equation}

Finally, to achieve uniform steps on the output distribution, we use \citeauthor{Pascanu}'s derivation of natural gradient given a loss function $\mathcal{L}$ \citeyearpar{Pascanu}:
\begin{equation}
\nabla \mathcal{L}_N = \nabla \mathcal{L} \mathbf{F_\theta}^{-1}
\end{equation} 

Using this definition and solving the Lagrange multiplier for minimizing the loss of parameters updated by $\Delta\theta$ under the constraint of a constant symmetric KL divergence, one can derive the approximation for constant symmetric KL divergence, using the information matrix. Taking the second-order Taylor expansion, \citet{Pascanu} derive:
\begin{equation}
KL_{\text{sym}}(p_\theta | p_{\theta+\Delta\theta})  \approx \frac{1}{2}\Delta\theta^T\mathbf{F}_\theta\Delta\theta
\end{equation}

As the output probability distribution is dependent on the final layer activation, \citet{Pascanu} give the following representation for a layer with a linear activation (interpreted as a conditional Gaussian distribution), here adapted for Q-learning, where $\beta$ is defined as the standard deviation: 
\begin{equation}
p_\theta(q|s)= \mathcal{N}(q | Q(s, \theta), \beta^2)
\end{equation}

In this formulation, since the information is only dependent on the final layer's activation we can use different activations in the hidden layers without changing the Fisher information. As in \citet{Pascanu}, the Fisher information can be derived where $\mathbf{J}_Q$ corresponds to the Jacobian of the output vector with respect to the parameters as follows:
\begin{equation} \label{eq:fisherinfo}
\mathbf{F}_{\text{linear}} = \beta^2 \mathbb{E}_{s\sim{d^{\pi}(s)}} \left[\mathbf{J}_Q^T\mathbf{J}_Q\right]
\end{equation}

\section{Related work}
We borrow heavily from the approach of \citet{Pascanu}, using their natural gradient for deep neural networks formalization and implementation in our method.

Next, we look at work on a different method of natural-gradient descent by \citet{Desjardins}. In this paper, algorithm called "Projected Natural Gradient Descent" (PRONG) is proposed, which also considers the Fisher information matrix in its derivation. While our paper does not explore this approach, it could be an area of future research, as PRONG is shown to converge better on multiple data-sets, such as CIFAR-10 \citep{Desjardins}.


Additional methods of applying natural gradient to reinforcement learning algorithms such as policy gradient and actor-critic are explored in \citet{Kakade} and \citet{Peters}. In both works, the natural variants of their respective algorithms are shown to perform favorably compared to their non-natural counterparts. Details on theory, implementation, and results are in their respective papers.


Insights into the mathematics of optimization using natural conjugate gradient techniques are provided in the work of \citet{Honkela}. These methods allow for more efficient optimization in high dimensions and nonlinear contexts.

The Natural Temporal Difference Learning algorithm applies natural gradient to reinforcement learning systems based on the Bellman error, although Q-learning is not explored \citep{naturalTD}. The authors use natural gradient with residual gradient, which minimizes the MSE of the Bellman error and apply natural gradient to SARSA, an on-policy learning algorithm. Empirical experiments show that natural gradient again outperforms standard methods in the tested environments.

Finally, to our knowledge, the only one other published or publicly available attempt of natural Q-learning was created by \citep{Barron}. In this work, the authors re-implemented PRONG and verified its efficacy at MNIST. However, when the authors tried to apply it to Q-learning, they got negative results, with no change on CartPole and worse results on GridWorld.

\section{Methods}
In our experiments, we use a standard method of Q-learning to act on the environment. Lasagne \citep{lasagne}, Theano \citep{theano}, and AgentNet \citep{yandex_2016} complete the brunt of the computational work. Because our implementation of natural gradient adapted from \citeauthor{Pascanu} originally fit an $X$ to a mapping $y$ and directly back-propagated a loss, we modify the training procedure to use a target value change similar to that described in equation \ref{qnetworkloss}. We also decay the learning rate by multiplying it by a constant factor every iteration.

As the output layer of our Q-network has a linear activation function, we use the parameterization of the Fisher information matrix for linear activations, which determines the natural gradient. For this, we refer to equation \ref{eq:fisherinfo}, approximated at every batch.

We calculate the desired change in parameters according to the Fisher information matrix as in \citet{Pascanu} by efficiently solving the system of linear equations relating the desired change in parameters to the gradients of the loss with respect to the weights: $Gx=\pdv{L}{\theta}$ (see Algorithm \ref{alg:ngdqn}). The MinRes-QLP Algorithm solves this linear equation by extending MinRes, an existing Krylov subspace descent algorithm to solve linear equations, to ill-conditioned systems such as a singular FIM using the QLP decomposition of the tridiagonal matrix from the Lanczos process \citep{minresqlp}. This method finds minimum length solutions robust to different conditions. We also test Linear Conjugate Gradient, an algorithm that solves the linear equation  by decomposing $x$ into vectors conjugate with respect to $G$ and iteratively calculating its components. Linear Conjugate Gradient is used for solving linear equations quickly and efficiently, with $\mathcal{O}(m\sqrt{k})$ where $m$ is the number of nonzero entries of $G$ and $k$ is its condition number \citep{lincg}.

For both algorithms, a damping factor $d$ is applied to ensure computability: $G:=G+d\textbf{I}$. Another efficiency of using linear solvers is that we are able to represent the FIM as an operator on $x$ without needing to explicitly compute the matrix. We take advantage of this by using Theano's left and right operator representations (Lop and Rop) of the Jacobian, as well as compare the linear solvers to Theano's explicit matrix inversion. This inversion utilizes the Gauss–Jordan elimination method to invert the Fisher information matrix with asymptotic time complexity $\mathcal{O}(n^3)$ \citep{theano}.

Our implementation runs on the OpenAI Gym platform which provides several classic control environments, such as the ones shown here, as well as other environments such as Atari \citep{gym}. The current algorithm takes a continuous space and maps it to a discrete set of actions.

In Algorithm \ref{alg:ngdqn}, we adapt \citeauthor{Mnih}'s Algorithm 1 and \citeauthor{Pascanu}'s Algorithm 2 \citeyearpar{Mnih, Pascanu}. Because these environments do not require preprocessing, we have omitted the preprocessing step, however this can easily be re-added. In our experiments, $\Delta\alpha$ was chosen somewhat arbitrarily to be $1-7e^{-5}$, and $\alpha$ was selected according to our grid-search (see: Hyperparameters). According to our grid search, we either leave the damping value unchanged or adjust it according to the Levenberg-Marquardt heuristic as used in \citet{Pascanu} and \citet{hessianfree}.

\begin{algorithm}
\begin{algorithmic}
\REQUIRE Initial learning Rate $\alpha_0$
\REQUIRE Learning rate decay $\Delta\alpha$
\REQUIRE Function \texttt{update\_damping}\\
\STATE Initialize replay memory $\mathcal{D}$ to capacity $N$\\
\STATE Initialize action-value function $Q$ with random weights\\
\STATE $\alpha \gets \alpha_0$\\
\FOR{$\text{episode}=1$, M}
\STATE Initialize sequence with initial state $s_1$ \\
\FOR{$t=1$, $T$}
\STATE With probability $\epsilon$ select a random action $a_t$, otherwise select action $a_t=\max_a Q^*(s_t, a;\theta)$ \\
\STATE Execute action $a_t$ in emulator and observe reward $r_t$ and state $s_{t+1}$\\
\STATE Store transition $(s_t, a_t, r_t, s_{t+1})$ in memory $\mathcal{D}$\\
\STATE Sample random minibatch of $n$ transitions $(s_j, a_j, r_j, s_{j+1})$ from $\mathcal{D}$\\
\STATE$y_j \gets \begin{cases}
r_j  \ifUSEICML\else&\fi \text{for terminal } s_{j+1} \\
r_j+\gamma \max_{a'}Q(s_{j+1},a';\theta) \ifUSEICML\else&\fi \text{for non-terminal } s_{j+1}
\end{cases}$\\
\STATE $g\gets\pdv{\mathcal{L}}{\theta}$ \\
\STATE $d\gets\texttt{update\_damping}(d)$\\
\STATE Define $G$ such that $G(v)=(\frac{1}{n}\textbf{J}_Q v)\textbf{J}_Q$\\
\STATE Solve $\text{argmin}_x\norm{(G+d\textbf{I})x-\pdv{L}{\theta}}$ with linear solver (e.g. MinresQLP \citep{minresqlp})\\
\STATE $\theta\gets\theta-\alpha x$ \\
\STATE $\alpha\gets\Delta_{\alpha}\alpha$
\ENDFOR
\ENDFOR
\caption{Natural Gradient Deep Q-Learning with Experience Replay}
\label{alg:ngdqn}
\end{algorithmic}
\end{algorithm}

\section{Experiments}
\subsection{Control Tasks}
To run Q-learning models on OpenAI gym, we adapt \citeauthor{Pascanu}'s implementation \citeyearpar{Pascanu}. For the baseline, we use OpenAI's open-source Baselines library \citep{baselines}, which allows reliable testing of tuned reinforcement learning architectures. As is defined in Gym, performance is measured by taking the best 100-episode reward over the course of running.

We run a grid search on the parameter spaces specified in the Hyperparameters section, measuring performance for all possible combinations. Because certain parameters like the exploration fraction are not used in our implementation of NGDQN, we grid search those parameters as well. As we wish to compare "vanilla" NGDQN to "vanilla" DQN, we test a version where target networks, model saving, or any other features, such as prioritized experience replay are not used. To further test the capabilities of NGDQN, we also compare NGDQN (which in these experiments are always run without target networkss) to DQN with target networks in order to show that the algorithm is competitive with other stabilization techniques.

\begin{figure*}[t!]
\subfloat[]{%
\begin{minipage}{0.501\textwidth} 
  \includegraphics[width=1\textwidth,keepaspectratio]{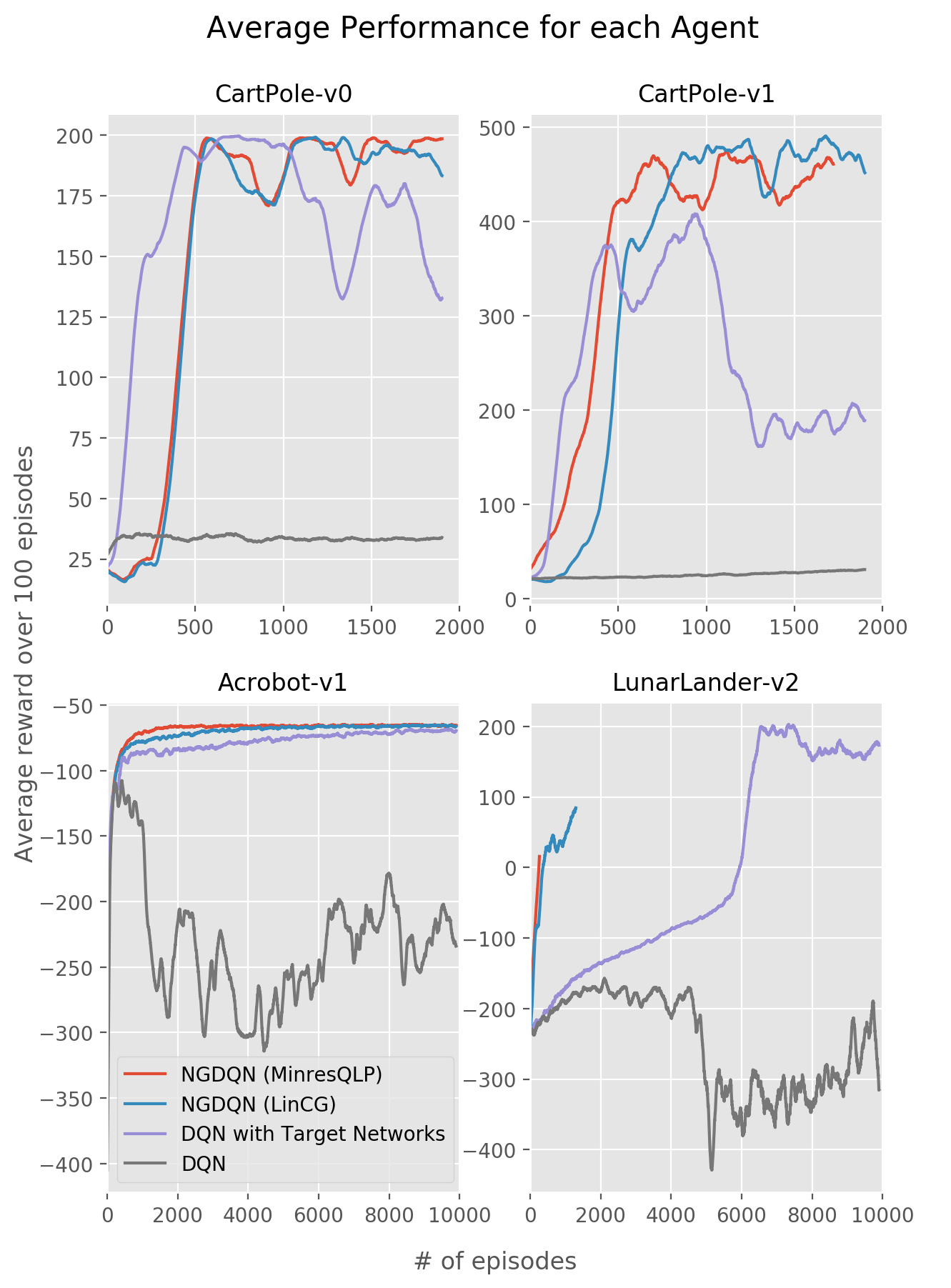}
\end{minipage}%
  }
    \subfloat[]{%
\begin{minipage}{0.5\textwidth}
  \includegraphics[width=1\textwidth,keepaspectratio]{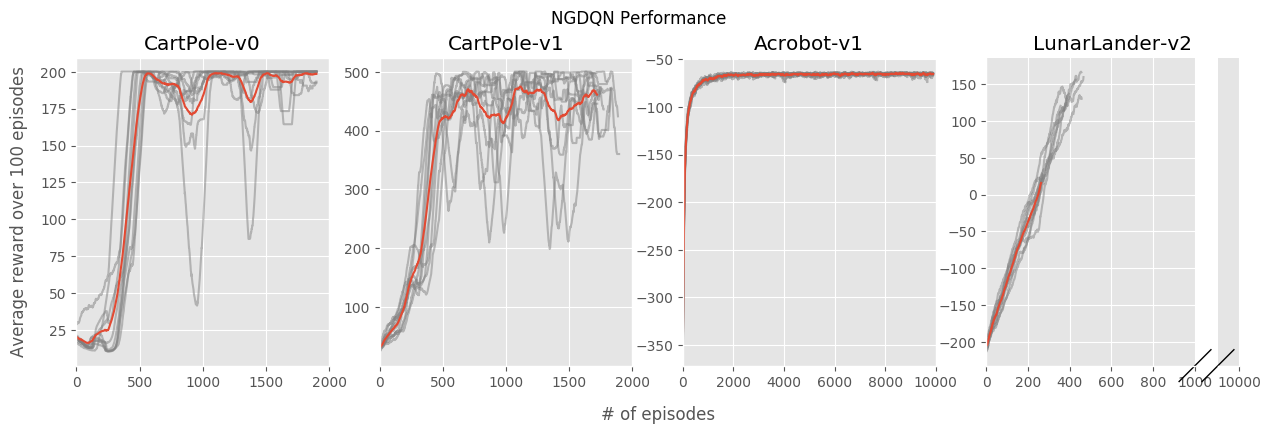}
  \includegraphics[width=1\textwidth,keepaspectratio]{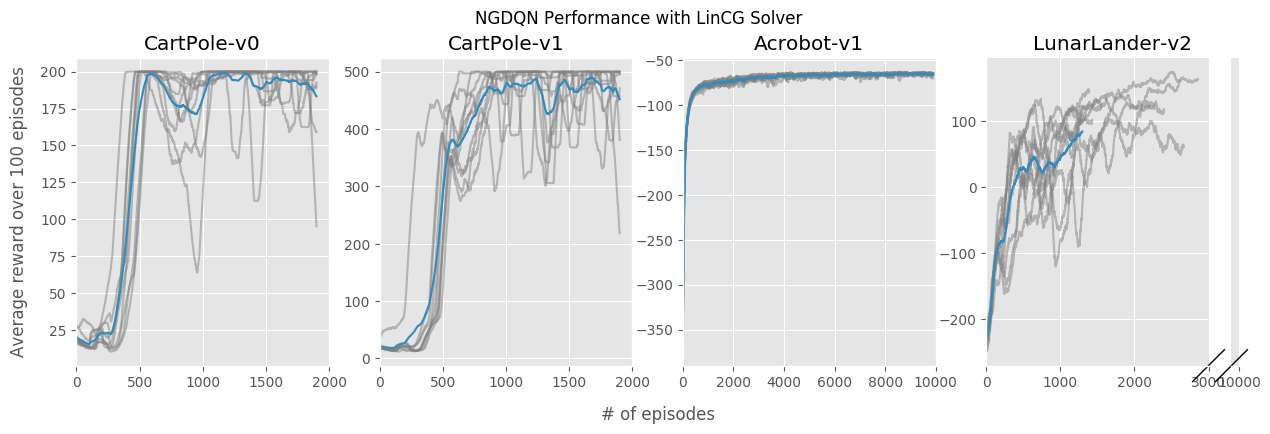}
  \includegraphics[width=1\textwidth,keepaspectratio]{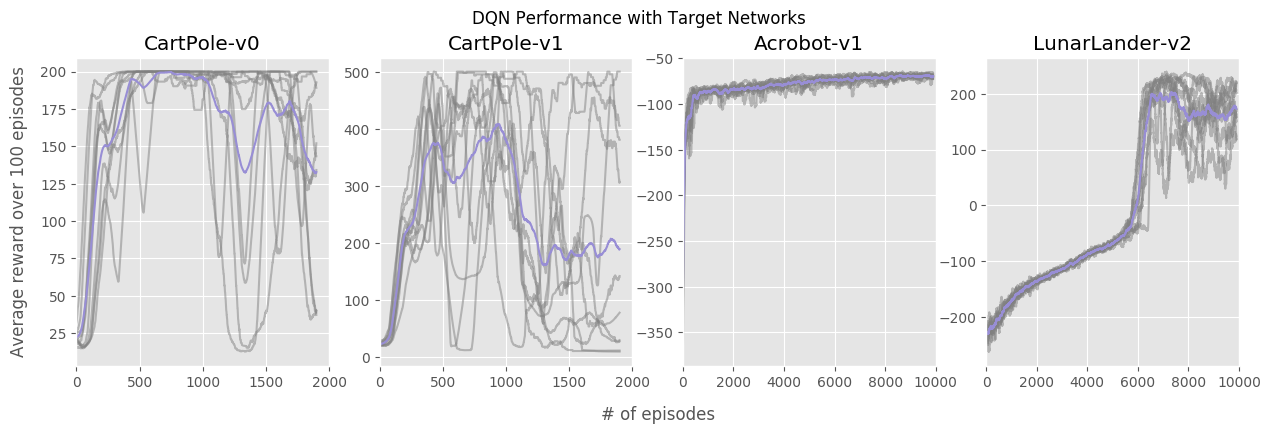}\hfill
  \includegraphics[width=1\textwidth,keepaspectratio]{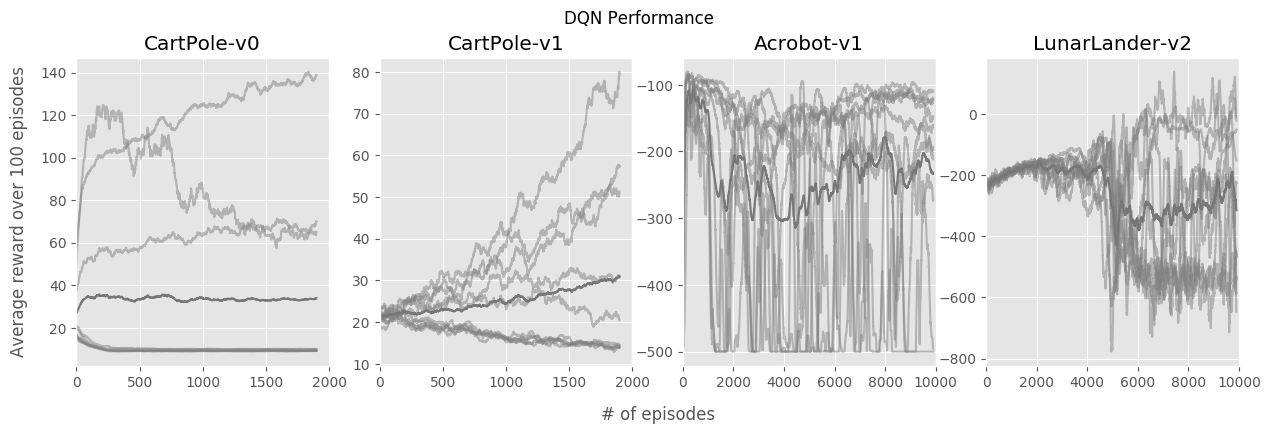}\hfill
  \end{minipage}
  }
 \caption{NGDQN and DQN performance over 10 trials over time with average line\protect\footref{lunarlander}. We can see that when training, NGDQN appears to be significantly more stable than the DQN baseline (i.e. NGDQN tended to reliably converge to a solution while the DQN baseline without target nests did not).}
\label{fig:graphs}
\end{figure*}

Following this grid search, we take the best result performance for each environment from both DQN and NGDQN, and run this configuration 10 times, recording a moving 100-episode average and the mean best 100-episode average across each run. These experiments reveal that NGDQN without target networks compare favorably to standard adaptive gradient techniques, even outperforming DQN with target networks. However, the increase in stability and speed comes with a trade-off: due to the additional computation, natural gradient takes longer to train when compared to adaptive methods, such as the Adam optimizer \citep{adam}. Details of this can be found in \citeauthor{Pascanu}'s work \citeyearpar{Pascanu}.

We test NGDQN in this manner on four common control environments from \url{https://github.com/openai/gym}: CartPole-v0, CartPole-v1, Acrobot-v1, and LunarLander-v2 (see Appendix B).\setcounter{footnote}{1}\footnotetext{\label{lunarlander}The LunarLander-v2 task for NGDQN was not completed, as the Stanford Sherlock cluster where the environments were run does not permit GPU tasks for over 48 hours. Therefore, each of the 10 trials was run for 48 hours and then stopped.}

\subsection{Inversion Methods}
During training, updates to the weights are calculated by solving a linear system (see Algorithm \ref{alg:ngdqn}), equivalent to matrix-vector-product of the inverse damped FIM with the gradient of the loss. We test the different methods to solve for these updates by computing the parameter updates from both MinRes-QLP and Linear Conjugate Gradient and comparing them to the updates given by an explicitly computed true FIM inversion. After separately solving for these individual updates, we calculate a variety of metrics to record how the natural gradient differs between inversion methods. This process is measured over 100 episodes of training on CartPole-v0, using the Linear CG's parameter updates.

For this NGDQN algorithm to satisfy the theoretical properties of natural gradient, an accurate inversion method is needed, and in order to create an effective yet efficient algorithm, it is necessary to balance accuracy and computational cost. Because natural gradient alters the step size of the gradient descent vector according to second order information and the angle of that vector through KL divergence, we record the norm of the calculated natural gradient and the angle between the update steps between true values solvers and estimators.\footnote{Calculated as $\arccos{(\hat{a}\cdot \hat{b})}$, where $\hat{a}$ is the flattened normalized updates computed by the true inverse and $\hat{b}$ is the flattened normalized updates given by the linear solvers.} We also record the computation time of each method. Finally, to ensure that damping is not significantly skewing the NG calculation, we compute the maximal eigenvalue of the Fisher information matrix by optimizing $\max_{\hat{x}} \hat{x}^TG\hat{x}$ (see Appendix C), which gives us an indication of the scaling done by the FIM. By comparing this value to the damping factor, we see that damping is relatively small and not significantly skewing the calculation of the natural gradient.

\section{Results}

\begin{figure*}
  \centering
  \includegraphics[width=0.9\textwidth,keepaspectratio]{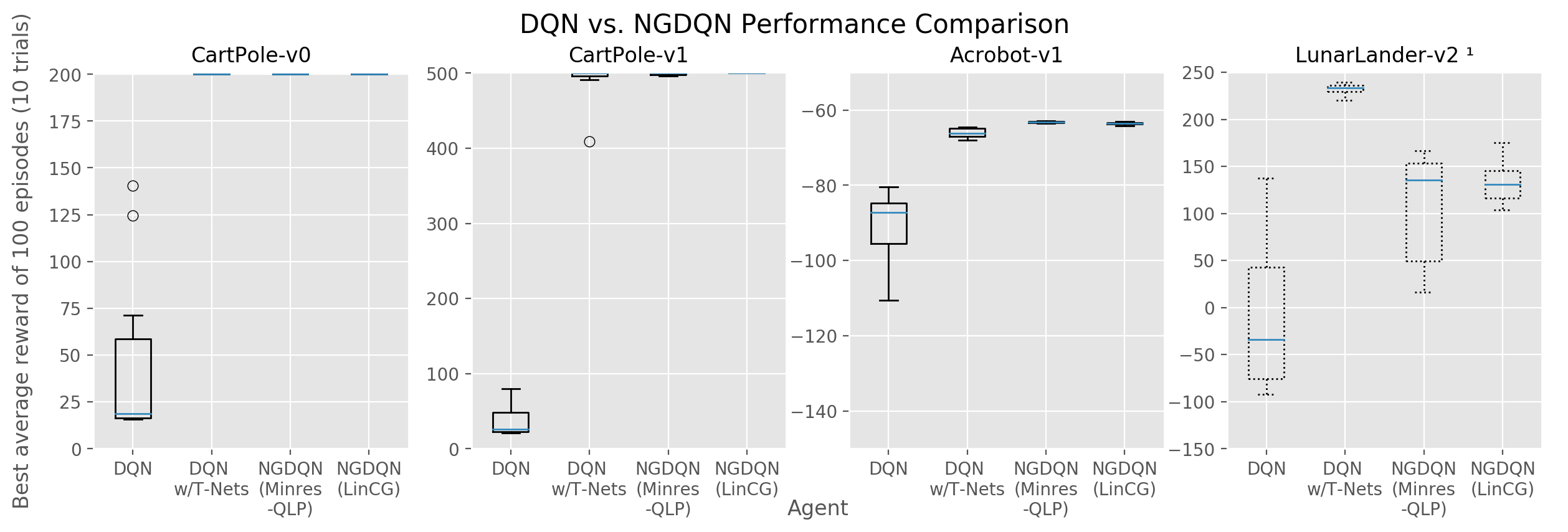}
  \caption{Average best 100-episode run over 10 trials with IQR. We can see that in every environment fully run, NGDQN achieves a higher max 100-episode average than our DQN baselines.\protect\footref{lunarlander}}
\end{figure*}

NGDQN and DQN were run against these four experiments, to achieve the following results summarized in Figure 1. The hyperparameters can be found in the Appendix A, and the code for this project can be found in the Appendix D. Each environment was run for a number of episodes (see Appendix A), and as per Gym standards the best 100 episode performance was taken.

In all experiments, natural gradient converges faster and achieves higher performance more consistently than the DQN benchmark, indicating its robustness in this task compared to the standard adaptive gradient optimizer used in the Baseline library (Adam). NGDQN also arrives at better solutions more reliably across the searched hyperparameters, exhibiting its versatility to different configurations when compared to the harsh tuning of DQN (see Appendix A). The success across all tests indicates that natural gradient generalizes well to diverse control tasks, from simpler tasks like CartPole to more complex tasks like LunarLander.

Comparison of the different inversion methods reveals that they calculate similar parameter updates. We find that MinRes-QLP and Linear CG arrive at updates with slightly smaller magnitudes and extremely similar directions as computing the true matrix inversion of the FIM, indicating that even with these approximations, the algorithm is  consistent with the theory behind natural gradient. It is also shown that the damping factor is most often between $5\%$ to $10\%$ of the maximal eigenvalue, indicating that the FIM is not over-damped. Finally, the compute times for the estimated FIM inversions are shown to be significantly less than that of the true FIM inversions, showing that using these methods helps accelerate training.

\section{Discussion}
\begin{figure}
    \centering
    \includegraphics[width=1\textwidth]{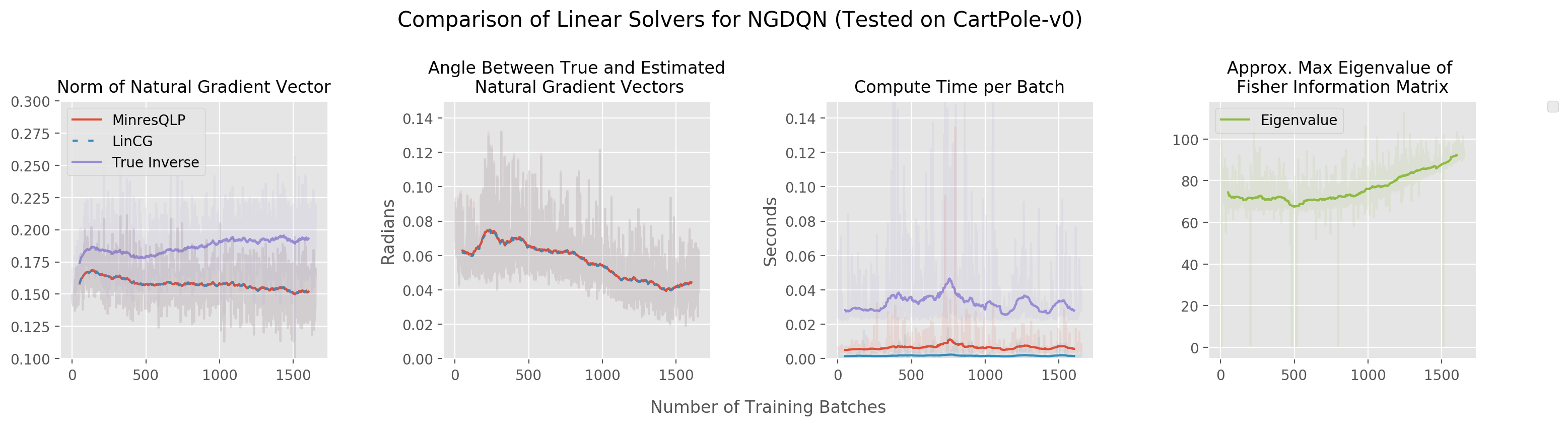}
    \caption[]{Summary of comparisons between different inversion methods. Moving averages of 100-batches are provided for readability, marked with solid lines.\footref{computetime}}
    \label{fig:inversions}
\end{figure}
In this paper, natural-gradient methods are shown to accelerate and stabilize training for common control tasks, even without target networks. This could indicate that Q-learning's instability may be diminished by naturally optimizing it, and also that natural gradient could be applied to other areas of reinforcement learning in order to address important problems such as sample efficiency.
\setcounter{footnote}{3}\footnotetext{\label{computetime}Tested on separate computer, hence slightly different compute time from agent training.} 

Although we have not yet empirically investigated the precise cause of this increase in stability, we offer a possible explanation below. One potential cause is that, although the replay buffer partially decorrelates the training set and time, the replay buffer will, over the course of training, become more filled with transitions from later on in each environment's episode. 

When playing, our agent's buffer will, during the beginning of training, be primarily filled with transitions from when the agent was acting poorly (e.g. in LunarLander-v2, the replay buffer is filled with transitions where the craft eventually plummets to a fiery death). However, much later in training when the agent has learned a policy to achieve higher rewards, the overall composition of this training buffer will be shifted to become more relevant to what the agent has to learn later in training. As the first few steps of gradient descent have a disproportionately large impact on the trained model, training with SGD could potentially be destabilized later on by these largely random transitions \cite{Pascanu}. 

In this scenario, target networks could help stabilize training. Natural gradient, by comparison, is very robust to reordering of the training set \citep{Pascanu}. This means that NGDQN could potentially use experience acquired later in training more effectively, as the overall policy of the agent would not be as skewed to experience gained early during training. This is, of course, only one possible explanation, and we hope that researchers will further investigate this phenomena in later work. 

\section*{Contributions \& Acknowledgements}
\ifBLIND
(Omitted for double-blind review.)
\fi
\ifNOTBLIND
Here, a brief contributor statement is provided, as recommended by \citet{winnerscurse}. 

Primary author led and directed the research, wrote the paper, edited the paper, wrote/adapted all the learning code for this project, and ran the RL experiments. Secondary author wrote and performed the inversion experiments, wrote parts of the paper, debugged and verified all of the code for correctness, and was important in editing the paper.

Thanks to Jen Selby for providing valuable insight and support, and for reviewing the paper and offering her suggestions over the course of the writing process. Thanks to Leonard Pon for his instruction, advice, and generous encouragement, especially during Applied Math, where the first part of this project took place. Thanks to Joshua Achiam for the valuable discussions, and for imparting his support, advice, and know-how. Also, huge thanks to Kavosh Asadi for providing us with valuable guidance, feedback, edits, and for helping us navigate the research scene. 
\fi

\newpage
\bibliographystyle{plainnat}
\bibliography{bib}
\newpage
\section*{Appendix A: Hyperparameters}
Both NGDQN and DQN had a minimum epsilon of $0.02$ and had a $\gamma$ of $1.0$ (both default for Baselines). The NGDQN model was tested using an initial learning rate of $1.0$. For NGDQN, the epsilon decay was set to $0.995$, but since there wasn't an equivalent value for the Baselines library, the grid search for Baselines included an exploration fraction (defined as the fraction of entire training period over which the exploration rate is annealed) of either $0.01$, $0.1$, or $0.5$ (see Table \ref{tab:baseline}). Likewise, to give baselines the best chance of beating NGDQN, we also searched a wide range of learning rates, given below.

\begin{table}[H]
\begin{center}
\begin{tabular}{lll}
\toprule
Environment    & \thead{\# of Episodes Ran For}& \thead{Layer \\ Configuration} \\ \midrule
CartPole-v0    & 2000                 & {[}64{]}\\
CartPole-v1    & 2000                 & {[}64{]}\\
Acrobot-v1     & 10,000               & {[}64, 64{]}\\
LunarLander-v2 & 10,000               & {[}256, 128{]}\\ \bottomrule
\end{tabular}
\end{center}
\caption{Shared configuration}
\end{table}

The batch job running time is given below (hours:minutes:seconds) for Sherlock. NGDQN LunarLander-v2 was run on the \texttt{gpu} partition which supplied either an Nvidia GTX Titan Black or an Nvidia Tesla GPU. All other environments were run on the \texttt{normal} partition. Additional details about natural gradient computation time can be found in \citet{Pascanu}.

\begin{table}[H]
\begin{center}
\begin{tabular}{lll}
\toprule
Environment    & \thead{NGDQN Batch Time}& \thead{DQN Batch Time} \\ \midrule
CartPole-v0    & 4:00:00    & 1:00:00 \\
CartPole-v1    & 9:00:00 & 1:00:00 \\
Acrobot-v1     & 48:00:00 & 8:00:00 \\
LunarLander-v2 & 48:00:00\footnotemark & 12:00:00 \\ \bottomrule
\end{tabular}
\end{center}
\caption{Running time}
\end{table}
\footnotetext{Jobs not completed; see Figure \ref{fig:graphs} for details}

Hyperparameter grid-search space:


\begin{table}[H]
\begin{center}
\begin{tabular}{@{}ll@{}}
\toprule
Hyperparameter  & Search Space           \\ \midrule
Learning Rate   & {[}0.01, 0.1, 1.0{]}\\
Adapt Damping   & {[}Yes, No{]}\\
Batch Size      & {[}32, 128{]}          \\
Memory Length   & {[}2500, 50000{]} \\
Activation & {[}Tanh, ReLU{]}           \\
\bottomrule
\end{tabular} 
\end{center}
\caption{NGDQN hyperparameter search space}
\end{table}

\begin{table}[H]
\begin{center}
\begin{tabular}{ll}
\toprule
{Hyperparameter}  & {Search Space}\\ \midrule
Learning Rate (no TNs)  & \makecell{{[}1e-08, 1e-07, 1e-06, 5e-06, 1e-05, \\5e-05, 0.0001, 0.0005, 0.005, 0.05{]}}\\
Learning Rate (with TNs)\footnotemark  & \makecell{{[}1e-08, 1e-07, 1e-06, 1e-05, 1e-04, 1e-03{]}}\\
Exploration Fraction            & {[}0.01, 0.1, 0.5{]}         \\
Batch Size      & {[}32, 128{]}          \\
Memory Length   & {[}500, 2500, 50,000{]} \\
Activation      & {[}Tanh, ReLU{]}           \\
Target Network Update Freq & {[}N/A, 500, 1000, 10,000{]} \\
\bottomrule
\end{tabular}
\end{center}
\caption{Baseline DQN hyperparameter search space}
\label{tab:baseline}
\end{table}

Best grid-searched configurations, used for experiments:\footnotetext{Due to training idiosyncrasies, the learning rate search space was different and some configurations for Acrobot-v1 were not run, although we believe given the results, this minor difference is insignificant}

\begin{table}[H]
\begin{center}
\begin{tabular}{l|lllll}
\toprule
Environment & \thead{Learning\\Rate} & \thead{Exploration\\ fraction} & \thead{Batch\\ Size} & \thead{Memory\\ Length} & \thead{Activ-\\ation} \\
\midrule
CartPole-v0     & 1e-07 & 0.01 & 128 & 2500   & Tanh \\
CartPole-v1     & 1e-08 & 0.1  & 32 & 50,000  & Tanh \\
Acrobot-v1      & 1e-05 & 0.01 & 128 & 50,000 & ReLU \\
LunarLander-v2  & 1e-05 & 0.01 & 128 & 2500   & Tanh \\ 
\bottomrule
\end{tabular}
\end{center}
\caption{Baseline DQN hyperparameter configuration}
\end{table}

\begin{table}[H]
\begin{center}
\begin{tabular}{l|llllll}
\toprule
Environment & \thead{Learning\\Rate} & \thead{Exploration\\ Fraction} & \thead{Batch\\ Size} & \thead{Memory\\ Length} & \thead{Target Net\\Update Freq} & \thead{Activ-\\ation}  \\
\midrule
CartPole-v0     & 0.001 & 0.01 & 128 & 50,000 & 500 & Tanh \\
CartPole-v1     & 0.001 & 0.01 & 32 & 50,000  & 500 & Tanh \\
Acrobot-v1      & 0.001 & 0.01 & 32 & 50,000 & 500 & Tanh \\
LunarLander-v2  & 0.0001 & 0.1 & 128 & 50,000  & 10,000 & ReLU \\ 
\bottomrule
\end{tabular}
\end{center}
\caption{Baseline DQN with target nets hyperparameter configuration}
\end{table}

\begin{table}[H]
\begin{center}
\begin{tabular}{l|lllll}
\toprule
Environment & \thead{Learning\\Rate} & \thead{Adapt\\Damping} & \thead{Batch\\ Size} & \thead{Memory\\ Length} & \thead{Activ-\\ation}  \\
\midrule
CartPole-v0    & 0.01 & No  & 128 & 50,000   & Tanh \\
CartPole-v1    & 0.01 & Yes & 128 & 50,000   & Tanh \\
Acrobot-v1     & 1.0  & No  & 128 & 50,000   & Tanh \\
LunarLander-v2 & 0.01 & No  & 128 & 50,000   & ReLU \\
\bottomrule
\end{tabular}
\end{center}
\caption{NGDQN (MinresQLP) hyperparameter configuration}
\end{table}

\begin{table}[H]
\begin{center}
\begin{tabular}{l|lllll}
\toprule
Environment & \thead{Learning\\Rate} & \thead{Adapt\\Damping} & \thead{Batch\\ Size} & \thead{Memory\\ Length} & \thead{Activ-\\ation}  \\
\midrule
CartPole-v0    & 0.01 & N/A  & 128  & 50,000  & Tanh \\
CartPole-v1    & 0.01 & N/A  & 128  & 50,000  & Tanh \\
Acrobot-v1     & 0.1  & N/A  & 128  & 50,000  & Tanh \\
LunarLander-v2 & 0.01 & N/A  & 128  & 50,000  & Tanh \\
\bottomrule
\end{tabular}
\end{center}
\caption{NGDQN (LinCG) hyperparameter configuration \protect\footnotemark}
\end{table}
\footnotetext{Damping adapt, learning rate $1.0$, memory length $500$, batch size $32$ not tested for LinCG due to poor results from initial tests to reduce computation burden}
\begin{figure*}[t!]
\subfloat[]{%
\begin{minipage}{0.5\textwidth}
  \includegraphics[width=1\textwidth,keepaspectratio]{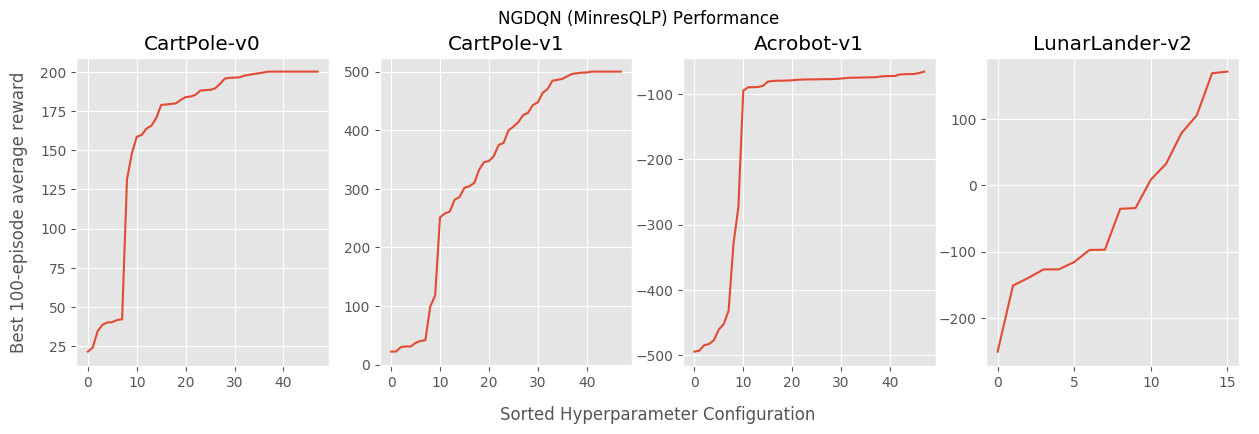}
  \includegraphics[width=1\textwidth,keepaspectratio]{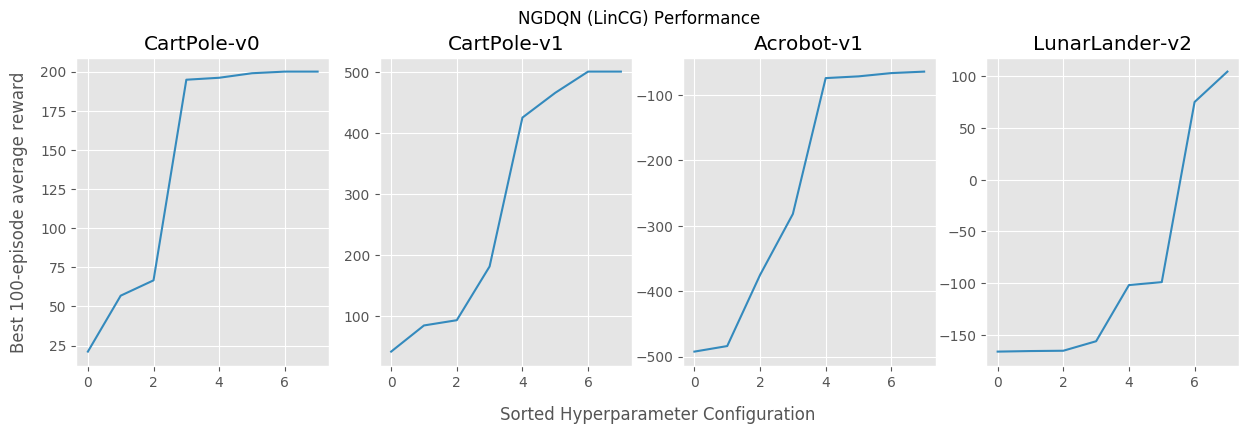}
  \end{minipage}
 }
    \subfloat[]{%
\begin{minipage}{0.5\textwidth}
  \includegraphics[width=1\textwidth,keepaspectratio]{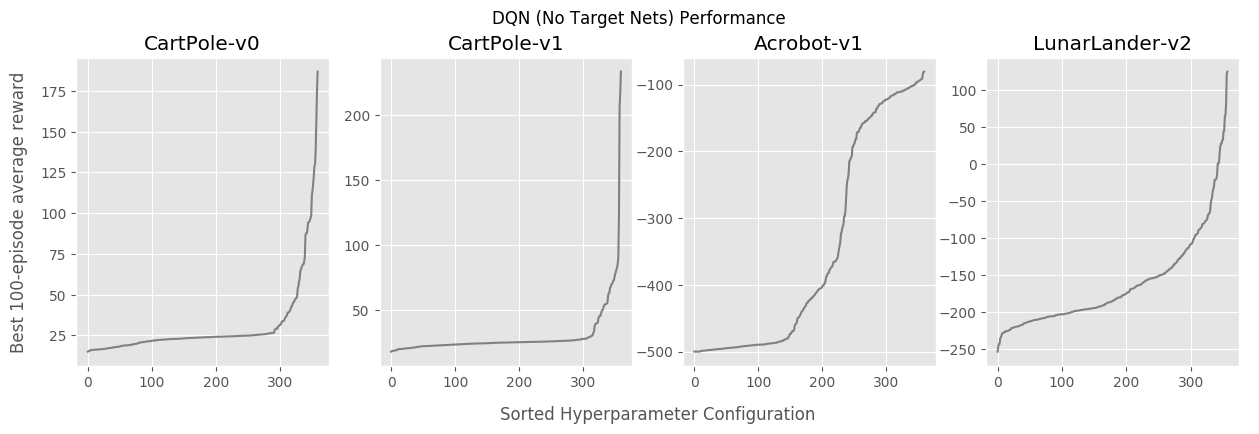}
  \includegraphics[width=1\textwidth,keepaspectratio]{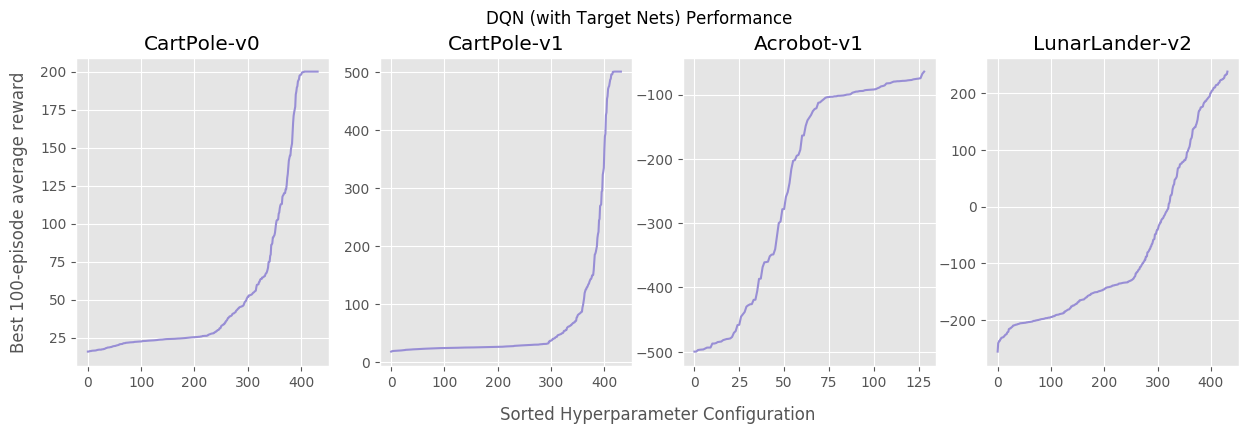}
  \end{minipage}
 }
 \caption{Grid searched performances over tested hyperparameter configurations for NGDQN (left) and DQN (right), ordered by increasing performance, hinting at robustness to changing hyperparameters}
\label{fig:hyper}
\end{figure*}
\section*{Appendix B: Environments}
Data is summarized from \url{https://github.com/openai/gym} and information provided on the wiki: \url{https://github.com/openai/gym/wiki}.
 
\subsection*{CartPole-v0}
The classic control task CartPole involves balancing a pole on a controllable sliding cart on a friction-less rail for 200 timesteps. The agent "solves" the environment when the average reward over 100 episodes is equal to or greater than 195. However, for the sake of consistency, we measure performance by taking the best 100-episode average reward.

The agent is assigned a reward for each timestep where the pole angle is less than $\pm12$ deg, and the cart position is less than $\pm2.4$ units off the center. The agent is given a continuous 4-dimensional space describing the environment, and can respond by returning one of two values, pushing the cart either right or left.

\subsection*{CartPole-v1}
CartPole-v1 is a more challenging environment which requires the agent to balance a pole on a cart for 500 timesteps rather than 200. The agent solves the environment when it gets an average reward of 450 or more over the course of 100 timesteps. However, again for the sake of consistency, we again measure performance by taking the best 100-episode average reward. This environment essentially behaves identically to CartPole-v0, except that the cart can balance for 500 timesteps instead of 200.

\subsection*{Acrobot-v1}
In the Acrobot environment, the agent is given rewards for swinging a double-jointed pendulum up from a stationary position. The agent can actuate the second joint by returning one of three actions, corresponding to left, right, or no torque. The agent is given a six dimensional vector describing the environments angles and velocities. The episode ends when the end of the second pole is more than the length of a pole above the base. For each timestep that the agent does not reach this state, it is given a $-1$ reward. 

\subsection*{LunarLander-v2}
Finally, in the LunarLander environment, the agent attempts to land a lander on a particular location on a simulated 2D world. If the lander hits the ground going too fast, the lander will explode, or if the lander runs out of fuel, the lander will plummet toward the surface. The agent is given a continuous vector describing the state, and can turn its engine on or off. The landing pad is placed in the center of the screen, and if the lander lands on the pad, it is given reward. The agent also receives a variable amount of reward when coming to rest, or contacting the ground with a leg. The agent loses a small amount of reward by firing the engine, and loses a large amount of reward if it crashes. Although this environment also defines a solve point, we use the same metric as above to measure performance.
\section*{Appendix C: Computation of the Maximal Eigenvalue}
To ensure our inversion is not overdamped, we compare the maximal eigenvalue the Fisher information matrix to its damping factor. To calculate this eigenvalue, we optimize $\max_x \hat{x}^TG\hat{x}$, as our FIM is implemented as a matrix-vector product only. Here we give pseudocode to calculate the eigenvector and outline a proof to show that this method has a global at minimum the maximal eigenvalue.

\begin{algorithm}
\begin{algorithmic}
\REQUIRE Matrix vector product $Gx$ given $x$
\REQUIRE Starting vector $v$, initialized randomly
\REQUIRE Early stopping condition ($c$: $0.001$)
\REQUIRE Training steps (steps: $10000$)
\REQUIRE Learning rate ($\alpha$: $0.0005$)
\FOR{$i=1$, steps}
\STATE $\Delta v \gets \nabla_v \left(\frac{v}{\left\Vert v\right\Vert } \cdot G\left(\frac{v}{\left\Vert v\right\Vert }^T\right)\right)$
\STATE $v \gets v + \alpha \Delta v$
\IF{$\left\Vert\Delta v\right\Vert_2 > c$} 
\STATE \textbf{break}
\ENDIF
\ENDFOR
\RETURN $\frac{v}{\left\Vert v\right\Vert } \cdot G\left(\frac{v}{\left\Vert v\right\Vert }^T\right)$
\caption{Algorithm to find the approximate maximum eigenvalue}
\label{alg:eigen}
\end{algorithmic}
\end{algorithm}

To prove this is maximized at the maximal eigenvalue of $G$, we show that $\hat{x}^TG\hat{x}$ is equivalent to the dot product $\hat{x} \cdot G\hat{x}$, which is also expressed as $\left\Vert \hat{x}\right\Vert \left\Vert G\hat{x}\right\Vert \cos{\theta}$ (where $\theta$ is the angle between $\hat{x}$ and $G\hat{x}$).

Clearly, $\left\Vert \hat{x}\right\Vert = 1$. Then, $\left\Vert Gx\right\Vert $ is maximized at the maximal eigenvalue of $G$. This is because any $\hat{x}$ can be decomposed into $\sum_{i=1}^{N_\lambda}c_i \hat{v}_i$ where $\hat{v}_i$'s are the unit eigenvectors of $G$. When $\hat{x}$ is operated by $G$,

\begin{equation}
    G\hat{x} = G\sum_{i=1}^{N_\lambda}c_i \hat{v}_i  = \sum_{i=1}^{N_\lambda}c_i G\hat{v}_i = \sum_{i=1}^{N_\lambda} \lambda_i c_i\hat{v}_i
\end{equation}

Thus, $\left\Vert G\hat{x}\right\Vert $ is maximized at $\left\Vert G\hat{x}\right\Vert=\lambda_{\text{max}}$ when $\hat{x}=\hat{v}_{\text{max}}$. Furthermore, $\cos{\theta}$ is maximized at $\theta=0$, which is also true when $\hat{x}=\hat{v}_{\text{max}}$ since $\hat{v}_{\text{max}}$ and $G\hat{v}_{\text{max}}$ are in the same direction by the definition of an eigenvector. Since each of the factors are maximized at this point, we thus show that the expression $\hat{x}^TG\hat{x}$ is maximized when $\hat{x}=\hat{v}_{\text{max}}$. It's maximal value is then

\begin{equation}
    \hat{v}_{\text{max}}^TG\hat{v}_{\text{max}} = \hat{v}_{\text{max}}^T(\lambda_\text{max}\hat{v}_{\text{max}}) = \lambda_\text{max}(\hat{v}_{\text{max}}^T\hat{v}_{\text{max}}) = \lambda_\text{max}
\end{equation}
\section*{Appendix D: Code}
\ifBLIND
The code for this project can be found at \url{https://github.com/ngdqn-authors/natural-gradient-deep-q-learning}. It uses a fork of OpenAI Baselines to allow for different activation functions: \url{https://github.com/ngdqn-authors/baselines}.
\fi
\ifNOTBLIND
The code for this project can be found at \url{https://github.com/hyperdo/natural-gradient-deep-q-learning}. It uses a fork of OpenAI Baselines to allow for different activation functions: \url{https://github.com/hyperdo/baselines}.
\fi
\end{document}